\documentclass{scrartcl}
\usepackage{enumitem}
\usepackage{mathtools}

\usepackage{amsmath}
\usepackage{amsfonts}
\usepackage{amssymb}
\usepackage{dsfont}
\usepackage[T1]{fontenc}
\usepackage[latin1]{inputenc}
\usepackage{epsfig}
\usepackage{graphicx}

\usepackage[boxed, lined]{algorithm2e}
\usepackage{float}
\usepackage{comment}

\newcommand{\bw}{\mathbf{w}}

\newcommand{\D}{{\mathcal{D}}}

\newcommand{\Nu}{{\mathcal{N}}}

\newcommand{\N}{\mathbb{N}}

\newcommand{\R}{\mathbb{R}}
\newcommand{\Z}{\mathbb{Z}}
\newcommand{\Rd}{\mathbb{R}^d}

\newcommand{\beq}{\begin{eqnarray*}}

\newcommand{\eeq}{\end{eqnarray*}}

\newcommand{\beqm}{\begin{eqnarray}}

\newcommand{\eeqm}{\end{eqnarray}}

\newtheorem{theorem}{Theorem}

\newtheorem{lemma}{Lemma}

\newcommand{\EXP}{{\mathbf E}}
\newcommand{\PROB}{{\mathbf P}}

\renewcommand{\bf}{\normalfont \bfseries}
\renewcommand{\it}{\normalfont \itshape}

\allowdisplaybreaks
\begin{document}
\renewcommand{\thefootnote}{\fnsymbol{footnote}}
\newcommand{\F}{{\cal F}}
\newcommand{\Sp}{{\cal S}}
\newcommand{\G}{{\cal G}}
\newcommand{\HH}{{\cal H}}

\begin{center}

  {\LARGE \bf
    On the rate of convergence of a deep recurrent neural network
    estimate in a regression problem with dependent data
  }
\footnote{
Running title: {\it Recurrent neural network estimates}}
\vspace{0.5cm}

Michael Kohler$^{1}$
and
Adam Krzy\.zak$^{2,}$\footnote{Corresponding author. Tel:
  +1-514-848-2424 ext. 3007, Fax:+1-514-848-2830}
\\

{\it $^1$
Fachbereich Mathematik, Technische Universit\"at Darmstadt,
Schlossgartenstr. 7, 64289 Darmstadt, Germany,
email: kohler@mathematik.tu-darmstadt.de}

{\it $^2$ Department of Computer Science and Software Engineering, Concordia University, 1455 De Maisonneuve Blvd. West, Montreal, Quebec, Canada H3G 1M8, email: krzyzak@cs.concordia.ca}

\end{center}
\vspace{0.5cm}

\begin{center}
October 25, 2020
\end{center}
\vspace{0.5cm}

\noindent
    {\bf Abstract}\\
    A regression problem with dependent data is considered.
    Regularity assumptions on the
    dependency of the data are introduced, and it is shown
    that under suitable structural assumptions on the
    regression function a deep recurrent neural network
    estimate is able to circumvent the curse of dimensionality.
    \vspace*{0.2cm}


\vspace*{0.2cm}


\section{Introduction}
\label{se1}

\subsection{Scope of this paper}
Motivated by the huge success of deep neural networks
in applications (see, e.g.,  Schmidhuber (2015),
Rawat and Wang (2017),
Hewamalage, Bergmeir and Bandara (2020)
and the literature cited therein) there is nowadays a
strong interest in showing theoretical properties
of such estimates. In the last years many new results
concerning deep feedforward neural network estimates
have been derived (cf., e.g.,
Eldan and Shamir (2016),
Lu et al. (2020),
Yarotsky (2018) and
Yarotsky and Zhevnerchuk (2019)
concerning approximation properties or
 Kohler and Krzy\.zak (2017), Bauer and Kohler
 (2019) and Schmidt-Hieber (2020) concerning statistical
 properties of these estimates).
 But basically no theoretical convergence results are known about
 the recurrent neural network estimates, which are among those
 neural network estimates which have been successfully
 applied in practice for time series forecasting (Smyl (2020), Mas and Carre (2020) and
 Makridakis, Spiliotis and Assimakopolulos (2018)),
 handwriting recognition (Graves  et al. (2008),
 Graves and Schmidhuber (2009)),
 speech recognition (Graves and Schmidhuber (2005),
 Graves, Mohamed and Hinton (2013)) and natural language processing
 (Pennington, Socher and Manning (2014)). For survey of the recent advances on
 recurrent neural networks see Salehinejad et al. (2018). In this paper
 we introduce a special class of deep recurrent neural
 network estimates and analyze their statistical properties
 in the context of  regression estimation problem with dependent data.

\subsection{A regression problem with dependent data}
 In order to motivate our regression estimation problem with dependent
 data, we start by considering
a general time series prediction problem with exogeneous variables
described
as follows: Let $(X_t,Y_t)$ $(t \in \Z)$ be $\Rd \times \R$--valued
random variables which satisfy
\begin{equation}
\label{se1eq1}
Y_t = F(X_t, (X_{t-1},Y_{t-1}), (X_{t-2},Y_{t-2}), \dots) + \epsilon_t
\end{equation}
for some measurable function
$F: \Rd \times (\Rd \times \R)^\N \rightarrow \R$ and
 some real-valued random variables $\epsilon_t$ with the property
\begin{equation}
\label{se1eq2}
\EXP\{\epsilon_t|X_t, (X_{t-1},Y_{t-1}), (X_{t-2},Y_{t-2}), \dots\}=0
\quad a.s.,
\end{equation}
where $\R$ and $\N$ are real numbers and positive integers, respectively.
 Given the data
\begin{equation}
\label{se1eq3}
\D_n=\{(X_1,Y_1), \dots, (X_n,Y_n)\}
\end{equation}
 the aim is to construct an estimate $m_n(\cdot)=m_n(\cdot,\D_n):\Rd
 \rightarrow \R$ such that the mean squared prediction error
\[
\EXP \left\{
\left|
Y_{n+1}-m_n(X_{n+1},\D_n)
\right|^2
\right\}
\]
is as small as possible.

In this article we simplify the above general model by imposing five main
constraints.
Firstly, we assume that
$F(X_t, (X_{t-1},Y_{t-1}), (X_{t-2},Y_{t-2}), \dots) $
does not depend on the complete infinite past
$(X_{t-1},Y_{t-1}), (X_{t-2},Y_{t-2}), \dots$
but only on the last $k$ times steps, where $k \in \N$.
Secondly, we assume that  $F(X_t, (X_{t-1},Y_{t-1}), (X_{t-2},Y_{t-2}), \dots) $
depends only on the $x$-values.
Thirdly, we assume that $F$ has, in addition,
a special recursive structure:
\begin{eqnarray*}
&&
F(X_t, (X_{t-1},Y_{t-1}), (X_{t-2},Y_{t-2}), \dots)
=
G(X_t, H_k(X_{t-1}, X_{t-2}, \dots, X_{t-k}))
\end{eqnarray*}
where
\begin{eqnarray}
&&
H_k(x_{t-1}, x_{t-2}, \dots, x_{t-k})
=
H(x_{t-1}, H_{k-1}(x_{t-2}, x_{t-3}, \dots, x_{t-k}))
\label{se1eq10}
\end{eqnarray}
and
\begin{equation}
  \label{se1eq11}
H_1(x_{t-1})=H(x_{t-1},0).
\end{equation}
Here $G:\Rd \times \R \rightarrow \R$ and  $H:\Rd \times \R
\rightarrow \R$ are smooth functions.
Fourthly, we assume that $\epsilon_t$ are independent
and identically distributed
random
variables with mean zero satisfying the following sub-Gaussian
assumption:
\begin{equation}
  \label{se1eq12}
\EXP \left\{
e^{c_1 \cdot \epsilon_t^2}
\right\}
<
\infty.
\end{equation}
And finally we simplify our model further by assuming
that $X_1$, $X_2$, \dots are independent and identically distributed.

In this way we get the following regression problem: Let $(X_t)_{t \in \Z}$
be independent identically distributed random variables with values in $\Rd$ and
let $(\epsilon_t)_{t \in \Z}$
be independent identically distributed random variables with values in $\R$, which are
independent of  $(X_t)_{t \in \Z}$.
Assume $\EXP \{\epsilon_t\}=0$ and (\ref{se1eq12}).
Set
\[
Y_t = G(X_t, H_k(X_{t-1}, \dots, X_{t-k})) + \epsilon_t
\]
for some (measurable) $G: \Rd \times \R \rightarrow \R$ and $H_k$ defined by
(\ref{se1eq10}) and (\ref{se1eq11}) for some (measurable)
$H:\Rd \times \R \rightarrow \R$.
Given the data (\ref{se1eq3})
we want to construct an estimate
\[
m_n(\cdot)=m_n(\cdot, \D_n): \Rd \times (\Rd)^k \rightarrow \R
\]
such that
\[
\EXP \left\{
\left|
Y_{n+1}-m_n(X_{n+1},X_n, \dots, X_{n-(k-1)})
\right|^2
\right\}
\]
is as small as possible.

In the above model we have
\[
\EXP\{Y_t | X_t=x_t, \dots, X_{t-k}=x_{t-k}\}
=
G( x_t, H_k(x_{t-1}, \dots, x_{t-k})),
\]
i.e.,
\begin{equation}
  \label{se1eq*1}
  m(x_1,\dots,x_{k+1})=G(x_{k+1},H_k(x_k,\dots,x_1))
\end{equation}
is the regression function on
\[
((X_1, \dots, X_{k+1}),Y_{k+1}),
\]
and our estimation problem above is a standard regression
estimation problem where we try to estimate (\ref{se1eq*1})
from the data
\begin{equation}
  \label{se1eq*2}
  ((X_1, \dots, X_{k+1}),Y_{k+1}),
  ((X_2, \dots, X_{k+2}),Y_{k+2}),
  \dots,
  ((X_{n-k}, \dots, X_{n}),Y_{n}).
  \end{equation}
Here the data (\ref{se1eq*2}) is not independent because
each of the variables $X_2$, \dots, $X_{n-1}$ occur in several of
the data ensembles.

\subsection{A recurrent neural network estimate}
We construct a recurrent neural network estimate
as follows: Below we define a suitable class $\F_n$ of recurrent
neural networks and use the least squares principle to define
our estimate by
\begin{equation}
\label{se1eq4}
\tilde{m}_n=
\arg \min_{f \in \F_n}
\frac{1}{n-k}
\sum_{t=k+1}^n
| Y_t
-
f(X_t,X_{t-1},  \dots, X_{t-k}) |^2
\end{equation}
and
\begin{equation}
\label{se1eq5}
m_n(X_{n+1}, \D_n)
=
T_{\beta_n} \tilde{m}_n(X_{n+1},X_{n},  \dots, X_{n-(k-1)} )
\end{equation}
where $T_L z=\min\{\max\{z,-L\},L\}$ (for $L >0$ and $z\in \R$)
is a truncation operator and $\beta_n= c_2 \cdot \log n$.

So it remains to define the class $\F_n$ of recurrent neural networks.
Here we use standard feedforward neural networks with additional
feedback loops.
We start by defining our artificial neural network
by choosing the so--called activation function $\sigma:\R \rightarrow
\R$, for which we select the ReLU activation function
\begin{equation}
  \label{se1eq9}
\sigma(z)=\max\{z,0\} \quad (z \in \R).
\end{equation}
Our neural network consists of $L$ layers of hidden neurons
with $k_l$ neurons in layer $l$. It
depends on  a vector of weights
$w_{i,j}^{(l)}$ and $\bar{w}_{j,(r,\bar{l})}^{(l)}$, where
$w_{i,j}^{(l)}$ is the weight between neuron $j$ in layer $l-1$ and
neuron
$i$ in layer $l$, and where
$\bar{w}_{j,(r,\bar{l})}^{(l)}$
is the recurrent weight between neuron $r$ in layer $\bar{l}$
and neuron $j$ in layer $l$.
For each neuron $j$ in layer $l$ the index set $I_j^{(l)}$ describes
the neurons in the neural network from which there exists
a recurrent connection to this neuron.
The function corresponding to this network
evaluated at $x_1,\dots,x_t$
is defined recursively as follows:
\begin{equation}
\label{se1eq6}
f_{net,\bw}(t)=
\sum_{j=1}^{k_L} w_{1,j}^{(L)} \cdot f_j^{(L)}(t),
\end{equation}
where
\begin{equation}
\label{se1eq7}
f_j^{(l)}(t)
=
\sigma
\left(
\sum_{s=1}^{k_{l-1}}
w_{j,s}^{(l)}
\cdot
f_s^{(l-1)}(t)
+
I_{\{t>1\}}
\cdot
\sum_{(r,\bar{l}) \in I_j^l}
\bar{w}_{j,(r,\bar{l})}^{(l)}
\cdot
f_r^{(\bar{l})}(t-1)
\right)
\end{equation}
for $l=2, \dots, L$ and
\begin{equation}
\label{se1eq8}
f_j^{(1)}(t)
=
\sigma
\left(
\sum_{s=0}^{d}
w_{j,s}^{(1)}
\cdot
x_t^{(s)}
+
I_{\{t>1\}}
\cdot
\sum_{(r,\bar{l}) \in I_j^1}
\bar{w}_{j,(r,\bar{l})}^{(1)}
\cdot
f_r^{(\bar{l})}(t-1)
\right).
\end{equation}
Here we have set $x_t^{(0)}=1$.
In case that $f_{net,\bw}$ is computed as above we
define
\[
f_{net,\bw}(X_t,X_{t-1},  \dots, X_{t-k})
\]
as $f_{net,\bw}(k+1)$ where the function is evaluated
at
$X_{t-k}$,
$X_{t-k+1}$,
\dots,
$X_t$. Here we set in  (\ref{se1eq8})
$x_{k+1}=X_t$, $x_k=X_{t-1}$, \dots, $x_1=X_{t-k}$.

\begin{figure}[!ht]
  \centering
      {
        \vspace*{-2cm}
\includegraphics[width=5cm]{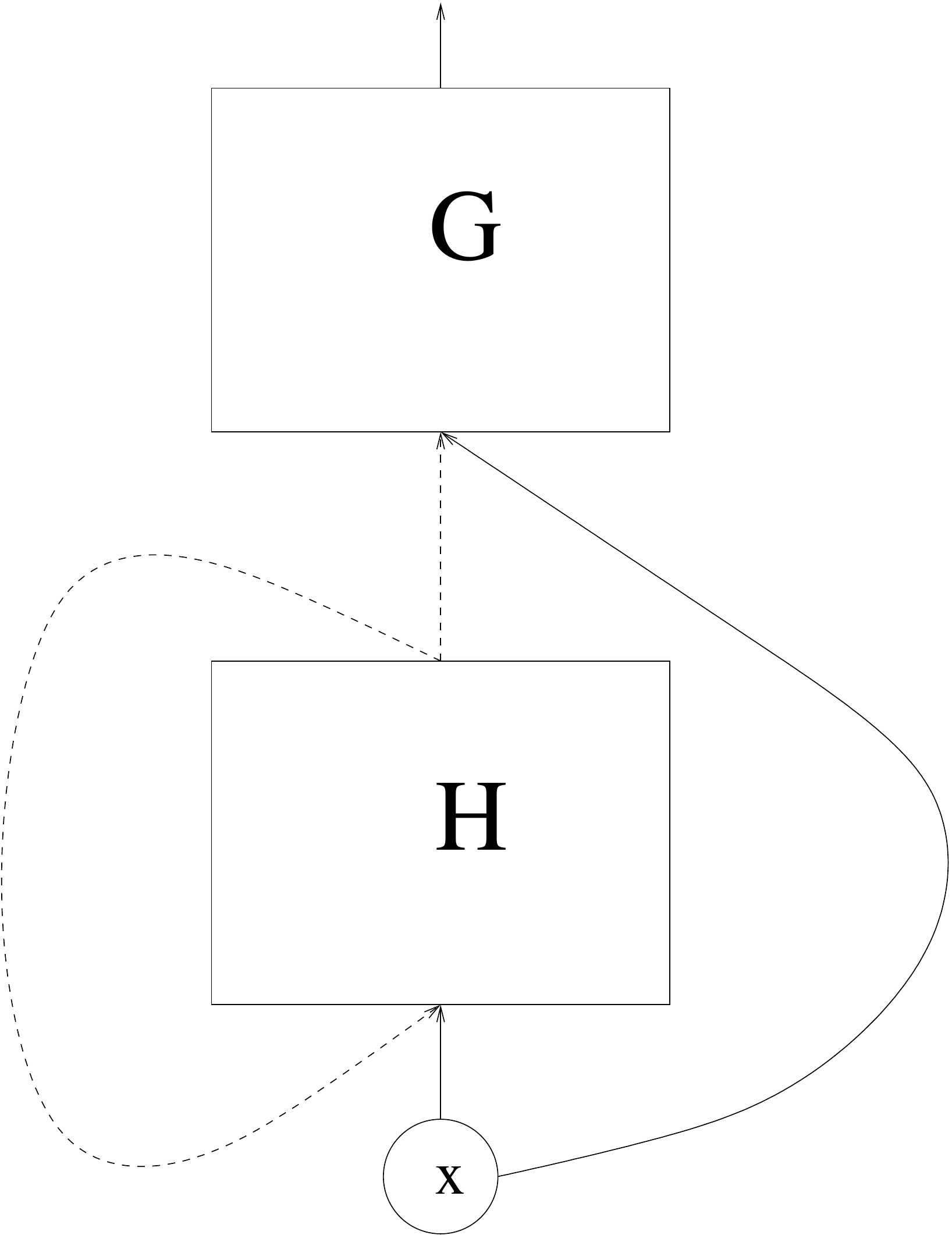}
}
\caption{\label{fig1}
  The structure of the recurrent neural networks. The solid
  arrows are standard feedforward connections, the dashed
  arrows represent the recurrent connections. The two boxes
  represent the parts of the network which approximate
   functions $G$ and $H$. Here $H$ is approximately computed in layers
  $1, \dots, L_{n,1}$, and function $G$ is approximately computed in layers $L_{n,1}+1, \dots, L_{n,1}+L_{n,2}$.
}
\end{figure}

In order to describe the above neural networks completely
(up to the weights, which are chosen in data-dependent way by the
least squares principle as described
in (\ref{se1eq4})) we have to choose the number $L$ of hidden
layers, the numbers of hidden neurons $k_1$, $k_2$, \dots, $k_L$
in layers $1$, $2$, \dots, $L$, and the location of the recurrent
connections described by the index sets $I_{j}^{(l)}$. We set
$L=L_{n,1}+L_{n,2}$, $k_l=K_{n,1}$ for $l=1, \dots, L_{n,1}$
and $k_l=K_{n,2}$ for $l=L_{n,1}+1$, \dots, $L_{n,1}+L_{n,2}$,
where $L_{n,1}$, $L_{n,2}$, $K_{n,1}$, $K_{n,2}$ are parameters
of the estimate chosen in Theorem \ref{th1} below.
The location of the recurrent connections is described
in Figure \ref{fig1}, which sketches
the architecture of the recurrent network
(see Section \ref{se2} for a formal
definition).

\subsection{Main result}
In Theorem \ref{th1} we show that the recurrent neural network
regression estimate
(\ref{se1eq4}) and (\ref{se1eq5}) with the above class
of recurrent neural networks satisfies in case that
$G$ and $H$ are $(p_G,C_G)$ and $(p_H,C_H)$--smooth the
error bound
\begin{eqnarray*}
  &&
  \hspace*{-0.3cm}
\EXP \left\{
\left|
Y_{n+1}-m_n(X_{n+1},X_n, \dots, X_{n-(k-1)})
\right|^2
\right\}
\\
&&
  \hspace*{-0.3cm}
\leq \min_{g: (\Rd)^{k+1} \rightarrow \R}
\EXP \left\{
\left|
Y_{n+1}-g(X_{n+1}, \dots, X_{n-(k-1)})
\right|^2
\right\}
+
c_{3} \cdot
(\log n)^6  \cdot
n^{- \frac{2 \cdot \min\{p_g,p_H\}}{2 \cdot \min\{p_g,p_H\}+(d+1)}}.
\end{eqnarray*}
Here the derived rate of convergence depends on $d+1$ and
not on the dimension \linebreak $(k+1) \cdot (d+1)$ of the predictors
in the data set (\ref{se1eq*2}). This shows that by using
recurrent neural networks it is possible to get under
the above assumptions on the structure of $H_k$ a rate of convergence
which does not depend on $k$ (and hence circumvents the
curse of dimensionality in this setting).

\subsection{Discussion of related results}

The Recurrent Neural Networks (RNN) are the class of artificial neural networks which can be described by the directed cyclic or acyclic graph and which exhibit temporal dynamic behaviour. Such networks can implement time delays and feedback loops. They are able to learn long-term dependencies from sequential and time-series data. In particular, properly trained RNN can model an arbitrary dynamical system.  The most popular architectures of RNN are Hopfield networks in which all connections are symmetric (Bruck (1990)), Bidirectional Associative Memory, that stores associative data as vectors (Kosko (1988)), Recursive Neural Networks in which the same set of weights are applied recursively over the structured input (Socher et al. (2011)), Long-Short Term Memory (LSTM), a network able to model long-term dependencies which has been very popular in natural language processing and speech recognition (Hochreiter and Schmidhuber (1997)) and is more robust to vanishing gradients than the classical RNN, and Gated Recurrent Units which are derived from RNN by adding gating units to them (Cho et al. (2014)) and which are more capable to learn long-term dependencies and are more robust to vanishing gradients than the classical RNN. Deep RNN have been surveyed by Schmidhuber (2015). Recent advances on RNN have been discussed in Salehinejad et al. (2018). The main problems with training RNNs by backpropagation are overfitting and vanishing gradients. Overfitting has generally been controlled by regularization, dropout, activation stabilization and hidden activation preservation, see Srivastava et al (2014) and Krueger et al. (2016). Theoretical analysis of RNNs learning has been lacking to date.

\subsection{Notation}
\label{se1sub7}
Throughout the paper, the following notation is used:
The sets of natural numbers, natural numbers including $0$,
integers
and real numbers
are denoted by $\N$, $\N_0$, $\Z$ and $\R$, respectively.
For $z \in \R$, we denote
the greatest integer smaller than or equal to $z$ by
$\lfloor z \rfloor$, and
$\lceil z \rceil$
is the smallest
integer greater than or equal to $z$.
Let $D \subseteq \R^d$ and let $f:\R^d \rightarrow \R$ be a real-valued
function defined on $\R^d$.
We write $x = \arg \min_{z \in D} f(z)$ if
$\min_{z \in \D} f(z)$ exists and if
$x$ satisfies
$x \in D$ and $f(x) = \min_{z \in \D} f(z)$.
For $f:\R^d \rightarrow \R$
\[
\|f\|_\infty = \sup_{x \in \R^d} |f(x)|
\]
is its supremum norm, and the supremum norm of $f$
on a set $A \subseteq \R^d$ is denoted by
\[
\|f\|_{\infty,A} = \sup_{x \in A} |f(x)|.
\]
Let $p=q+s$ for some $q \in \N_0$ and $0< s \leq 1$.
A function $f:\R^d \rightarrow \R$ is called
$(p,C)$-smooth, if for every $\alpha=(\alpha_1, \dots, \alpha_d) \in
\N_0^d$
with $\sum_{j=1}^d \alpha_j = q$ the partial derivative
$\frac{
\partial^q f
}{
\partial x_1^{\alpha_1}
\dots
\partial x_d^{\alpha_d}
}$
exists and satisfies
\[
\left|
\frac{
\partial^q f
}{
\partial x_1^{\alpha_1}
\dots
\partial x_d^{\alpha_d}
}
(x)
-
\frac{
\partial^q f
}{
\partial x_1^{\alpha_1}
\dots
\partial x_d^{\alpha_d}
}
(z)
\right|
\leq
C
\cdot
\| x-z \|^s
\]
for all $x,z \in \R^d$.

Let $\F$ be a set of functions $f:\Rd \rightarrow \R$,
let $x_1, \dots, x_n \in \Rd$ and set $x_1^n=(x_1,\dots,x_n)$.
A finite collection $f_1, \dots, f_N:\Rd \rightarrow \R$
  is called an $L_2$ $\varepsilon$--cover of $\F$ on $x_1^n$
  if for any $f \in \F$ there exists  $i \in \{1, \dots, N\}$
  such that
  \[
  \left(
  \frac{1}{n} \sum_{k=1}^n |f(x_k)-f_i(x_k)|^2
  \right)^{1/2}< \varepsilon.
  \]
  The $L_2$ $\varepsilon$--covering number of $\F$ on $x_1^n$
  is the  size $N$ of the smallest $L_2$ $\varepsilon$--cover
  of $\F$ on $x_1^n$ and is denoted by $\Nu_2(\varepsilon,\F,x_1^n)$.

For $z \in \R$ and $\beta>0$ we define
$T_\beta z = \max\{-\beta, \min\{\beta,z\}\}$. If $f:\R^d \rightarrow
\R$
is a function and $\F$ is a set of such functions, then we set
$
(T_{\beta} f)(x)=
T_{\beta} \left( f(x) \right)$.

\subsection{Outline of the paper}
\label{se1sub8}
In Section \ref{se2} the deep recurrent
neural network estimates used in this paper
are defined.
The main result is presented in Section \ref{se3} and proven
in Section \ref{se4}.

\section{A recurrent neural network estimate}
\label{se2}
\begin{figure}[!ht]
  \centering
      {
        \vspace*{-2cm}
\includegraphics[width=5cm]{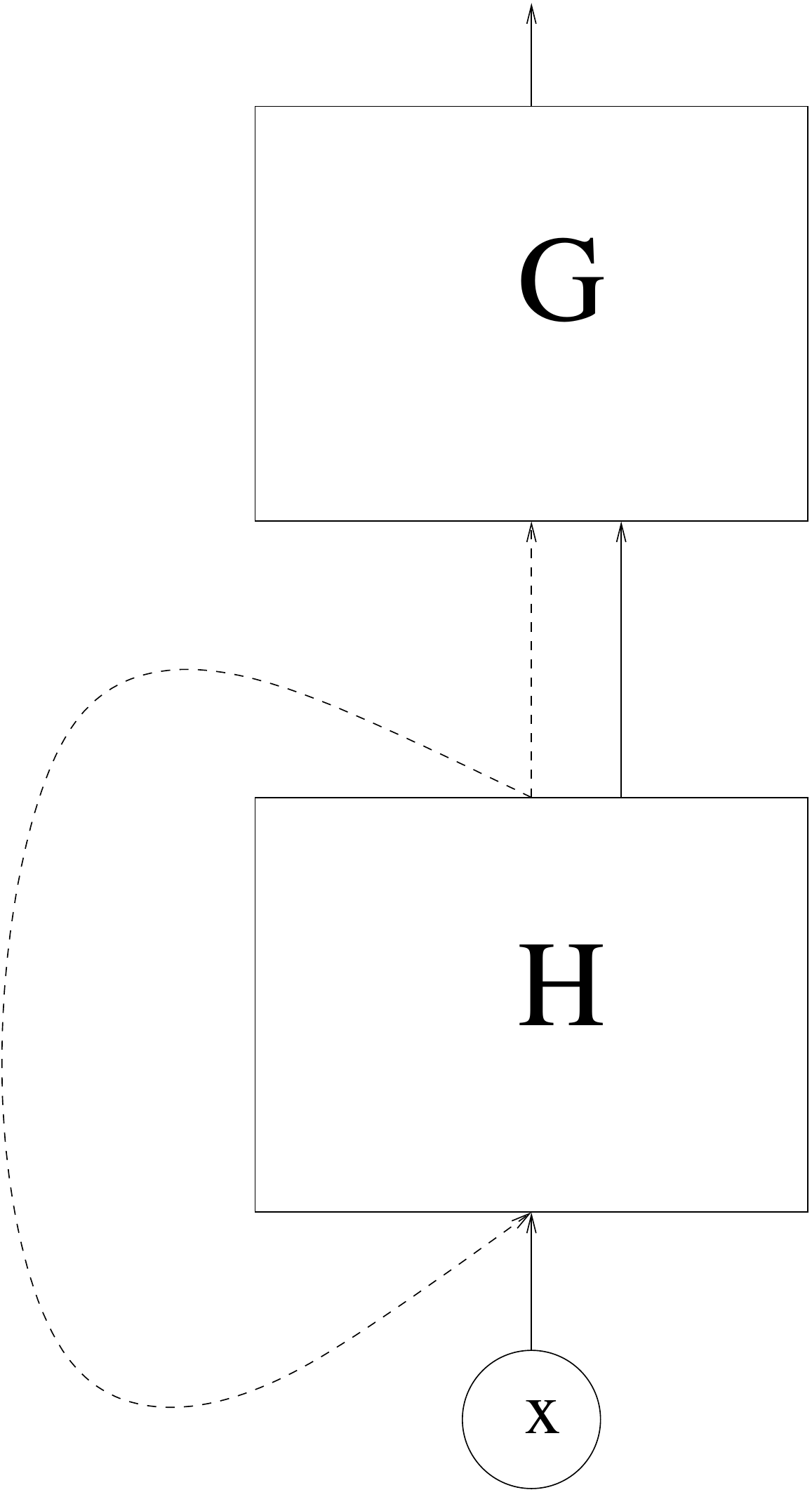}
}
\caption{\label{fig2}
  The structure of the recurrent neural networks in
  $\F(k,K_{n,1},K_{n,2},L_{n,1},L_{n,2})$. The solid
  arrows are standard feedforward connections, the dashed
  arrows represent the recurrent connections. The two boxes
  represent the parts of the network which implement approximations
  of functions $G$ and $H$. Here the network which approximates $H$
  also feeds the input to $G$.
}
\end{figure}

We start with the definition of our class
of the recurrent neural networks. It depends on parameters
$k$, $L_{n,1}$, $L_{n,2}$, $K_{n,1}$ and $K_{n,2}$.
As activation function
we use the ReLU activation function
defined in (\ref{se1eq9}). Depending on a weight vector $\bw$
which consists of weights
$w_{i,j}^{(l)}$ and $\bar{w}_{j,(r,\bar{l})}^{(l)}$ we define
our recurrent neural network
\[
f_{net,\bw}: (\Rd)^{k+1} \rightarrow \R
\]
by
\[
f_{net,\bw}(x_{k+1},x_{k}, \dots, x_1)
=
\sum_{j=1}^{K_{n,2}} w_{1,j}^{(L)} \cdot f_j^{(L_{n,1}+L_{n,2})}(k+1),
\]
where $ f_j^{(L_{n,1}+L_{n,2})}(t)$ are recursively defined as follows:
\begin{equation}
  \label{se2eq1}
f_j^{(l)}(t)
=
\sigma
\left(
\sum_{s=1}^{K_{n,2}}
w_{j,s}^{(l)}
\cdot
f_s^{(l-1)}(t)
\right)
\end{equation}
for $l=L_{n,1}+2, \dots, L_{n,1}+L_{n,2}$,
\begin{equation}
  \label{se2eq2}
f_j^{(L_{n,1}+1)}(t)
=
\sigma
\left(
\sum_{s=1}^{K_{n,1}}
w_{j,s}^{(L_{n,1}+1)}
\cdot
f_s^{(L_{n,1})}(t)
  +
I_{\{t>1\}}
\cdot
\sum_{s=1}^{K_{n,1}}
\bar{w}_{j,(s,L_{n,1})}^{(L_{n,1}+1)}
\cdot
f_s^{(L_{n,1})}(t-1)
\right),
  \end{equation}
\begin{equation}
  \label{se2eq3}
f_j^{(l)}(t)
=
\sigma
\left(
\sum_{s=1}^{K_{n,1}}
w_{j,s}^{(l)}
\cdot
f_s^{(l-1)}(t)
\right)
  \end{equation}
for $l=2, \dots, L_{n,1}$
and
\begin{equation}
  \label{se2eq4}
  f_j^{(1)}(t)
=
\sigma
\left(
\sum_{s=0}^{d}
w_{j,s}^{(1)}
\cdot
x_t^{(s)}
+
I_{\{t>1\}}
\cdot
\sum_{s=1}^{K_{n,1}}
\bar{w}_{j,(s,L_{n,1})}^{(1)}
\cdot
f_s^{(L_{n,1})}(t-1)
\right).
  \end{equation}
Let $\F(k,K_{n,1},K_{n,2},L_{n,1},L_{n,2})$ be the
class of all such recurrent deep networks. Observe
that here we implement the networks in a slightly different
way than in Figure \ref{fig1} since we do not use
a direct connection from the input to the part
of the network in G, instead we use the network which
implements $H$ also to feed the input to $G$ (cf., Figure \ref{fig2}).

Then our estimate is defined by
\begin{equation}
  \label{se2eq6}
\tilde{m}_n=
\arg \min_{f \in \F(k,K_{n,1},K_{n,2},L_{n,1},L_{n,2})}
\frac{1}{n-k}
\sum_{t=k+1}^n
| Y_t
-
f(X_t,X_{t-1},  \dots, X_{t-k}) |^2
\end{equation}
and
\begin{equation}
  \label{se2eq7}
m_n(X_{n+1}, \D_n)
=
T_{\beta_n} \tilde{m}_n(X_{n+1},X_{n},  \dots, X_{n-(k-1)} ).
  \end{equation}

\section{Main result}
\label{se3}
Our main result is described the following theorem.

	\begin{theorem}
	  \label{th1}
          Let $X_t$ $(t \in \Z)$ be independent
and identically distributed
          $[0,1]^d$--valued
          random variables, and let $\epsilon_t$ $(t \in \Z)$ be
          independent and identically distributed
          $\R$-valued random variables with
          $\EXP\{\epsilon_t\}=0$ which satisfy (\ref{se1eq12})
          and which are independent from $(X_t)_{t \in \Z}$.
          Let $G, H: \Rd \times \R \rightarrow \R$ be
          $(p_G,C_g)$-- and $(p_H,C_H)$--smooth functions which
          satisfy
          \begin{equation}
            \label{th1eq1}
            |G(x,z_1)-G(x,z_2)| \leq C \cdot |z_1-z_2|
            \quad \mbox{and} \quad
            |H(x,z_1)-H(x,z_2)| \leq C \cdot |z_1-z_2|
          \end{equation}
$            (x \in \Rd, z_1,z_2 \in \R)$
          for some constant $C >1$.
          Let $k \in \N$ and define
          \[
Y_t = G(X_t, H_k(X_{t-1},\dots,X_{t-k}))+\epsilon_t
\]
for $H_k$ recursively defined by (\ref{se1eq10}) and (\ref{se1eq11}).

Set
\[
K_{n,1}= \lceil c_4 \rceil, \quad K_{n,2}= \lceil c_5 \rceil,
\quad
 L_{n,1}=\left\lceil
c_6 \cdot n^{\frac{d+1}{2 \cdot (2 p_H + d+1)}} \right\rceil
\]
and
\[
 L_{n,2}=
\left\lceil
c_7 \cdot n^{\frac{d+1}{2 \cdot (2 p_G + d+1)}}
\right\rceil
\]
and define the estimate $m_n$ as in Section \ref{se2}.
Then we have for
$c_4$, \dots, $c_7>0$ sufficiently large and for
any $n \geq 2 \cdot k+2$:
\begin{eqnarray*}
  &&
  \hspace*{-0.3cm}
  \EXP \left\{
\left|
Y_{n+1}-m_n(X_{n+1},X_n, \dots, X_{n-(k-1)})
\right|^2
\right\}
\\
&&
  \hspace*{-0.3cm}
\leq \min_{g: (\Rd)^{k+1} \rightarrow \R}
\EXP \left\{
\left|
Y_{n+1}-g(X_{n+1}, \dots, X_{n-(k-1)})
\right|^2
\right\}
+
c_{8} \cdot
(\log n)^6  \cdot
n^{- \frac{2 \cdot \min\{p_G,p_H\}}{2 \cdot \min\{p_G,p_H\}+(d+1)}}.
  \end{eqnarray*}
	\end{theorem}

        \noindent
            {\bf Remark 1.}
            Our estimation problem can be considered as a regression
            problem with independent variable
            \[
(X_t,X_{t-1},\dots,X_{t-k}),
            \]
            having dimension $d \cdot k$.
            The rate of convergence in Theorem \ref{th1} corresponds
            to the optimal minimax rate of convergence of a regression problem
            with dimension $d+1$ (cf., Stone (1982)),
            hence our assumption on the
            structure of $H_t$ enables us to get the rate of convergence
            independent of $k$.

            \section{Proofs}
\label{se4}

\subsection{Auxiliary results from empirical process theory}
\label{se4sub1}
In our proof we will apply well-known techniques from the empirical
process theory as described, for instance, in van de Geer (2000).
We reformulate the results there by the following two
auxiliary lemmas.

Let
\[
Y_i = m(x_i) + W_i
\quad
(i=1, \dots, n)
\]
for some
$x_1, \dots, x_n \in \R^d$, $m:\R^d \rightarrow \R$ and some
random variables
$W_1$, \dots, $W_n$ which are independent and have expectation zero.
We assume that the $W_i$'s are sub-Gaussian in the sense that
\begin{equation}
\label{se52eq1}
\max_{i=1, \dots, n} K^2 \EXP \{ e^{W_i^2 / K^2} -1 \} \leq \sigma_0^2
\end{equation}
for some $K, \sigma_0 >0$.
Our goal is to estimate $m$ from
$(x_1, Y_1), \dots, (x_n, Y_n)$.
Let $\F_n$ be a set of functions $f:\R^d \rightarrow \R$
and consider the least squares estimate
\begin{equation}
\label{se52eq2}
\tilde{m}_n(\cdot)
=
\arg \min_{f \in \F_n}
\frac{1}{n} \sum_{i=1}^n |f(x_i)- Y_{i}|^2
\quad \mbox{and} \quad
m_n = T_{\beta_n} \tilde{m}_n,
\end{equation}
where
$\beta_n = c_2 \cdot \log n$.

\begin{lemma}
\label{le1}
Assume that the sub-Gaussian condition (\ref{se52eq1})
and
\[
|m(x_i)| \leq \beta_n/2 \quad (i=1, \dots,n)
\]
hold,
and let the estimate be defined by (\ref{se52eq2}). Then
there exist constants
$c_{9},c_{10}>0$
which depend only on $\sigma_0$ and $K$
such that for any
$\delta_n > c_{9} /n$ with
\begin{eqnarray}
\label{le1eq1}
&&
\sqrt{n} \cdot \delta
\geq
c_{9}
\int_{ \delta / (12 \sigma_0)}^{\sqrt{48 \delta}}
\Bigg(
\log \Nu_2 \Bigg(
u
,
\{ T_{\beta_n} f-g :  f  \in \F_n, \nonumber \\
&&
\hspace*{2.6cm}
\frac{1}{n} \sum_{i=1}^n
|T_{\beta_n} f(x_i)-g(x_i)|^2 \leq 4 \delta  \}
,
x_1^n
\Bigg)
\Bigg)^{1/2} du
\end{eqnarray}
for all $\delta \geq \delta_n / 6$
and all $g \in \F_n$
we have
\begin{eqnarray*}
&&
  \PROB \Bigg\{
  \frac{1}{n} \sum_{i=1}^n | m_n(x_i)-m(x_i)|^2
>
c_{10} \left(
\delta_n
+
\min_{f \in \F_n}
  \frac{1}{n} \sum_{i=1}^n | f(x_i)-m(x_i)|^2
\right)
\Bigg\}
\\
&&
\leq
c_{10} \cdot \exp \left(
- \frac{n \cdot \min \{ \delta_n, \sigma_0^2 \} }{c_{10}}
\right) + \frac{c_{10}}{n}.
\end{eqnarray*}
\end{lemma}
\noindent
{\bf Proof.}
Lemma \ref{le1} follows  from  the proof of
    proof of Lemma 3 in Kohler and Krzy\.zak (2020).
    For the sake of completeness a complete proof
    can be found in the Appendix.
\hfill $\Box$

In order to formulate our next auxiliary result we
let $(X,Y), (X_1,Y_1),\ldots$ be independent and identically
distributed $\R^d\times\R$ valued random variables with
$\EXP Y^2 <\infty$, and we let
$m(x) = \EXP\{Y|X=x\}$ be the corresponding
regression function.

\begin{lemma}
\label{le2}
Let $\beta_n \geq L \geq 1$ and assume that
 $m$ is bounded in absolute value by $L$. Let $n, N \in \N$, let
$\F_n$ be a set of functions $f:\R^d \rightarrow
\R$,
let
\[
\tilde{m}_n(\cdot)=\tilde{m}_n(\cdot, (X_1,Y_1),
\dots, (X_{n+N},Y_{n+N})) \in \F_n
\]
and set $m_n=T_{\beta_n} \tilde{m}_n$.
 Then
there exist constants $c_{11},c_{12},c_{13},c_{14}>0$
such that for any  $\delta_n>0$
which satisfies
\[
\delta_n > c_{11} \cdot \frac{\beta_n^2}{n}
\]
and
\begin{eqnarray}
  \label{le2eq1}
&&  c_{12} \cdot \frac{\sqrt{n} \delta}{\beta_n^2}
\geq
\int_{ c_{13} \cdot \delta / \beta_n^2
}^{\sqrt{\delta}} \Bigg( \log \Nu_2 \Bigg( u , \{
(T_{\beta_n} f-m)^2 \, : \, f \in \F_n
 \} , x_1^n \Bigg)
\Bigg)^{1/2} du
\end{eqnarray}
for all $\delta \geq \delta_n$ and all $x_1, \dots, x_n \in \R^d$,
 we have for $n \in \N \setminus \{1\}$
\begin{eqnarray*}
  &&
  \PROB \left\{
    \int |m_n(x)-m(x)|^2 \PROB_X(dx)
>
\delta_n
+
3
\frac{1}{n}
\sum_{i=1}^n
|m_n(X_i)-m(X_i)|^2
\right\}
\\
&&
\leq
c_{14} \cdot
\exp \left(
- \frac{n \cdot  \delta_n }{c_{14} \cdot \beta_n^2}
\right)
.
\end{eqnarray*}

\end{lemma}

\noindent
    {\bf Proof.} The result follows from the
    proof of Lemma 4 in Kohler and Krzy\.zak (2020).
    For the sake of completeness a complete proof
    can be found in the Appendix.
    \hfill $\Box$

\subsection{Approximation results for neural networks}

\begin{lemma}
  \label{le3}
  Let $d \in \N$,
  let $f:\Rd \rightarrow \R$ be $(p,C)$--smooth for some $p=q+s$,
  $q \in \N_0$  and $s \in (0,1]$, and $C>0$. Let $A \geq 1$
    and $M \in \N$ sufficiently large (independent of the size of $A$, but
     \begin{align*}
       M \geq 2 \ \mbox{and} \ M^{2p} \geq c_{15} \cdot \left(\max\left\{A, \|f\|_{C^q([-A,A]^d)}
       \right\}\right)^{4(q+1)},
     \end{align*}
     where
     \[
     \|f\|_{C^q([-A,A]^d)}
     =
     \max_{\alpha_1, \dots, \alpha_d \in \N_0, \atop \alpha_1 + \dots + \alpha_d \leq q}
     \left\|
\frac{
\partial^q f
}{
\partial x_1^{\alpha_1}
\dots
\partial x_d^{\alpha_d}
}
     \right\|_{\infty, [-A,A]^d}
     ,
     \]
 must hold for some sufficiently large constant $c_{15} \geq 1$).
 \\
a) Let $L, r \in \N$ be such that
\begin{enumerate}
\item $L \geq 5+\lceil \log_4(M^{2p})\rceil \cdot \left(\lceil \log_2(\max\{q, d\} + 1\})\rceil+1\right)$
\item $r \geq 2^d \cdot 64 \cdot \binom{d+q}{d} \cdot d^2 \cdot (q+1) \cdot M^d$
\end{enumerate}
hold.
There exists a feedforward neural network
$f_{net, wide}$ with ReLU activation function, $L$ hidden layers
and $r$ neurons per hidden layer such that
\begin{align}
 \| f-f_{net, wide}\|_{\infty, [-A,A]^d} \leq
  c_{16} \cdot \left(\max\left\{A, \|f\|_{C^q([-A,A]^d)}\right\} \right)^{4(q+1)} \cdot M^{-2p}.
  \label{le3eq1}
\end{align}
b) Let $L, r \in \N$ be such that
\begin{enumerate}
\item $L \geq 5M^d+\left\lceil \log_4\left(M^{2p+4 \cdot d \cdot (q+1)} \cdot e^{4 \cdot (q+1) \cdot (M^d-1)}\right)\right\rceil \\
  \hspace*{4cm}
\cdot \lceil \log_2(\max\{q,d\}+1)\rceil+\lceil \log_4(M^{2p})\rceil$
\item $r \geq 132 \cdot 2^d\cdot   \lceil e^d\rceil
  \cdot \binom{d+q}{d} \cdot \max\{ q+1, d^2\}$
\end{enumerate}
hold.
There exists a feedforward neural network
$f_{net, deep}$ with ReLU activation function, $L$ hidden layers
and $r$ neurons per hidden layer such that
 (\ref{le3eq1}) holds with
$f_{net,wide}$ replaced by $f_{net,deep}$.
  \end{lemma}
\noindent
    {\bf Proof.} See Theorem 2 in Kohler and Langer (2020).
    \hfill $\Box$

\begin{lemma}
  \label{le4}
  Let $k \in \N$, $x_1, \dots, x_{k+1} \in [0,1]^d$, $A \geq 1$,
  $g, \hat{g}:\Rd \times \R \rightarrow \R$,
  $h: \Rd \times \R \rightarrow [-A,A]$,
  $\hat{h}: \Rd \times \R \rightarrow \R$
  and assume
  \[
  |g(x,z) - g(x,\bar{z})| \leq C_{Lip,g} \cdot |z-\bar{z}|
  \quad \mbox{and} \quad
  |h(x,z) - h(x,\bar{z})| \leq C_{Lip,h} \cdot |z-\bar{z}|
  \]
  for some $ C_{Lip,g} ,  C_{Lip,h} >1$. Set $z_0=\hat{z}_0=0$,
  \[
  z_t = h(x_t,z_{t-1}) \quad \mbox{and} \quad
    \hat{z}_t = \hat{h}(x_t,\hat{z}_{t-1})
    \]
    for $t=1, \dots, k$. Assume
    \[
\frac{C_{Lip,h}^k-1}{C_{Lip,h}-1} \cdot \|h - \hat{h}\|_{\infty,[-2A,2A]^{d+1}} \leq 1.
    \]
    Then we have
    \begin{eqnarray*}
      &&
    |g(x_{k+1},z_k)
    -
    \hat{g}(x_{k+1}, \hat{z}_{k})|\\
    &&
    \leq
    \|g - \hat{g}\|_{\infty, [-2A,2A]^{d+1}} + C_{Lip,g} \cdot \frac{C_{Lip,h}^k-1}{C_{Lip,h}-1}
    \cdot     \|h - \hat{h}\|_{\infty, [-2A,2A]^{d+1}}.
    \end{eqnarray*}
\end{lemma}

\noindent
    {\bf Proof.}
    For $t \in \{1, \dots, k\}$, $z_{t-1} \in [-A,A]$  and
    $\hat{z}_{t-1} \in [-2A,2A]$  we have
    \begin{eqnarray*}
      |z_{t} - \hat{z}_{t}|
      &=&
      |      h(x_t,z_{t-1}) - \hat{h}(x_t,\hat{z}_{t-1})|
      \\
      &
      \leq&
            |h(x_t,z_{t-1}) - h(x_t,\hat{z}_{t-1})|
      +
      | h(x_t,\hat{z}_{t-1})- \hat{h}(x_t,\hat{z}_{t-1})|
      \\
      &\leq&
      C_{Lip,h} \cdot |z_{t-1} - \hat{z}_{t-1}|
      +
      \|h - \hat{h}\|_{\infty, [-2A,2A]^{d+1}}.
    \end{eqnarray*}
In case $z_s \in [-A,A]$  and
$\hat{z}_{s} \in [-2A,2A]$ for $s \in \{0,1,\dots,t-1\}$
we can conclude
\begin{eqnarray*}
&&      | z_{t} - \hat{z}_{t}| \\
      &&\leq
      \|h - \hat{h}\|_{\infty, [-2A,2A]^{d+1}} \cdot (1+ C_{Lip,h} +  C_{Lip,h}^2 +
      \dots +  C_{Lip,h}^{k-1}) + C_{Lip,h}^{k}\cdot |z_0-\hat{z}_0|
      \\
      &&=
      \|h - \hat{h}\|_{\infty, [-2A,2A]^{d+1}} \cdot
      \frac{C_{Lip,h}^k-1}{C_{Lip,h}-1}
      + 0
      \leq 1
\end{eqnarray*}
(where the last equality follows from $z_0=\hat{z}_0=0$),
which implies
\[
|\hat{z}_{t}|
\leq
|z_t|+
| z_{t} - \hat{z}_{t}|
\leq
A + 1 \leq 2A.
\]
Via induction we can conclude that
we have
$z_{s} \in [-A,A]$  and
$\hat{z}_{s} \in [-2A,2A]$
for $s \in \{0,1,\dots,k\}$ and consequently
we get
\[
    | z_{k} - \hat{z}_{k}|
      \leq
      \|h - \hat{h}\|_{\infty,[-2A,2A]^{d+1}} \cdot
      \frac{C_{Lip,h}^k-1}{C_{Lip,h}-1}.
\]
    This together with
    \begin{eqnarray*}
    |g(x_k,z_{k})
    -
    \hat{g}(x_k,\hat{z}_{k})|
    &\leq&
    |g(x_k,z_{k})
    -
    g(x_k,\hat{z}_{k})
    |
    +
    |g(x_k,\hat{z}_{k}) -
    \hat{g}(x_k,\hat{z}_{k})| \\
    & \leq &
    C_{Lip,g} \cdot |z_{k} - \hat{z}_{k}| + \| \hat{g}-g\|_{[\infty, -2A,2A]^d}
    \end{eqnarray*}
    implies the assertion.
    \hfill $\Box$

\begin{lemma}
  \label{le5}
  Let $k \in \N$ and $A \geq 1$.
  Assume that $g$ and $h$
are $(p_G,C_G)$-- and $(p_H,C_H)$--smooth functions which
  satisfy the assumptions
  of Lemma \ref{le4}, and define
  \[
  h_t(x_{t}, x_{t-1}, \dots, x_1) =
  h(x_t, h_{t-1}(x_{t-1}, x_{t-2},\dots, x_1))
  \]
  $t=2, \dots, k$ and
  \[
h_1(x_1)=h(x_1,0).
\]
Let $h_{net}$ be a feedforward neural network with
$L_{n,1}$ hidden layers and $K_{n,1}$ hidden neurons
in each layer and let
$g_{net}$ be a feedforward neural network with
$L_{n,2}$ hidden layers and $K_{n,2}$ hidden neurons
in each layer, which approximate $h$ and $g$.
Let $x_1, \dots, x_n \in [0,1]^d$ arbitrary and assume
\[
\|h_{net} - h\|_{\infty, [-2A,2A]^d} \cdot
\frac{C_{Lip,h}^k-1}{C_{Lip,h}-1}
\leq 1.
\]
Then there exists $f_{net,rec} \in \F(k,K_{n,1}+ 2 \cdot d,K_{n,2},L_{n,1},L_{n,2})$
such that
\begin{eqnarray*}
  &&
  | g(x_{k+1},h_k(x_k,\dots,x_1)) - f_{net,rec}(x_{k+1},\dots,x_1)|
  \\
  &&
\leq
c_{17} \cdot \max\{
\|g_{net} - g\|_{\infty, [-2A,2A]^{d+1}},
\|h_{net} - h\|_{\infty, [-2A,2A]^{d+1}}
\}
\end{eqnarray*}
holds for any $x_{k+1}, \dots, x_1 \in [0,1]^d$.
\end{lemma}

\noindent
    {\bf Proof.}
    We construct our recurrent neural network as follows:

    In layers $1, \dots, L_{n,1}$ it computes in neurons $1, \dots, K_{n,1}$
    $h_{net}(x,z)$, where $x$ is the input of the recurrent neural
    network and $z$ is the output of layer $L_{n,1}$ of the
    network in the previous time step propagated
    by the recurrent connections. In the same layer
    it uses
    \[
f_{id}(x)=x=\sigma(x)-\sigma(-x)
    \]
    in order to propagate in the neurons
    $K_{n,1}+1$, \dots, $K_{n,1}+2 \cdot d$
    the input value of $x$ to the next layer.

    In layers $L_{n,1}+1, \dots, L_{n,1}+L_{n,2}$ it computes
    in the neurons $1, \dots, K_{n,2}$ the function
    $g_{net}(x,z)$. Here the layer $L_{n,1}+1$
    gets as input the value of $x$ propagated to the layer $L_{n,1}$
    in the neurons $K_{n,1}+1$, \dots, $K_{n,1}+2 \cdot d$ in the
    previous layers, and (via a recurrent connection) the
    output $z$ of the network $h_{net}$ computed in the layers
    $1, \dots, L_{n,1}$ in the previous time step.

    The output of our recurrent network is the output of $g_{net}$
    computed in layer $L_{n,1}+L_{n,2}$.

    By construction, this recurrent neural network computes
    \[
    f_{net,rec}(x_{k+1},\dots,x_1)
    =
    g_{net}(x_{k+1},\hat{z}_k),
    \]
    where
    $\hat{z}_k$
    is recursively defined by
    \[
\hat{z}_t=h_{net}(x_t,\hat{z}_{t-1})
\]
for $t=2, \dots, k$ and
\[
\hat{z}_1=h_{net}(x_1,0).
\]
From this we get the assertion by applying Lemma \ref{le4}.
    \hfill $\Box$

\subsection{A bound on the covering number}

\begin{lemma}
  \label{le6}
Let $\F(k,K_{n,1},K_{n,2},L_{n,1},L_{n,2})$ be the
class of deep recurrent networks introduced
in Section \ref{se2} and assume
\[
\max\{L_{n,1},L_{n,2}\} \leq L_n \leq n^{c_{18}}
\quad \mbox{and} \quad
\max\{K_{n,1},K_{n,2}\} \leq K_n.
\]
Then we have for any
$z_1^s \in ((\Rd)^{k+1})^s$ and any $1/n^{c_{19}} < \epsilon < c_2 \cdot (\log n) / 4$
\begin{eqnarray*}
  &&
  \log \left(
  \Nu_2 \left(\epsilon,
  \{ T_{\beta_n} f \, : \, f \in \F(k,K_{n,1},K_{n,2},L_{n,1},L_{n,2}) \},
  z_1^s
  \right)
  \right)
  \\
  &&
  \leq
    c_{20} \cdot k \cdot L_n^2 \cdot K_n^2 \cdot (\log n)^2.
  \end{eqnarray*}
\end{lemma}

\noindent
    {\bf Proof.} By unfolding the recurrent neural networks
    in $\F(k,K_{n,1},K_{n,2},L_{n,1},L_{n,2})$ in time it is easy
    to see that $\F(k,K_{n,1},K_{n,2},L_{n,1},L_{n,2})$ is contained
    in a class of standard feedforward neural networks with
    \[
    (k+1) \cdot (L_{n,1}+L_{n,2})
    \]
    layers having at most
    \[
    \max\{K_{n,1},K_{n,2}\}+2d+2
    \]
    neurons per layer. In this
    unfolded feedforward neural network there are at most
    \[
    c_{22} \cdot (L_{n,1} \cdot K_{n,1}^2 + L_{n,2} \cdot K_{n,2}^2)
    \]
    different weights (since we share the same weights at all time
    points). By Theorem 6 in Bartlett et al. (2019) we
    can conclude that the VC dimension of the set
    of all subgraphs from $\F(k,K_{n,1},K_{n,2},L_{n,1},L_{n,2})$
    (cf., e.g., Definition 9.6 in Gy\"orfi et al. (2002))
    and hence also the VC dimension
    of the set of all subgraphs
    from
    \[
    \{ T_{\beta_n} f \, : \, f \in \F(k,K_{n,1},K_{n,2},L_{n,1},L_{n,2}) \}
    \]
    is bounded above by
    \begin{eqnarray*}
      &&
    c_{22} \cdot (L_{n,1} \cdot K_{n,1}^2 + L_{n,2} \cdot K_{n,2}^2)
    \cdot (k+1) \cdot
    (L_{n,1} + L_{n,2}) \cdot \log( (k+1) \cdot (L_{n,1} + L_{n,2}))
    \\
    &&
    \leq
    c_{23} \cdot k \cdot L_n^2 \cdot K_n^2 \cdot \log(n).
    \end{eqnarray*}
    From this together with Lemma 9.2 and Theorem 9.4
    in Gy\"orfi et al. (2002) we can conclude
    \begin{eqnarray*}
      &&
  \Nu_2 \left(\epsilon,
  \{ T_{\beta_n} f \, : \, f \in \F(k,K_{n,1},K_{n,2},L_{n,1},L_{n,2}) \},
  z_1^k
  \right)
  \\
  &&
  \leq
  3 \cdot \left(
\frac{4 e \cdot (c_{2} \cdot \log n)^2}{\epsilon^2}
\cdot
\log
\frac{6 e \cdot (c_{2} \cdot \log n)^2}{\epsilon^2}
\right)^{    c_{23} \cdot k \cdot L_n^2 \cdot K_n^2 \cdot \log(n)
},
      \end{eqnarray*}
    which implies the assertion. \hfill $\Box$

\subsection{Proof of Theorem \ref{th1}}
           {\it In the first step of the proof} we show that the
           assertion follows from
           \begin{eqnarray}
             \label{pth1eq1}
             &&
             \EXP \int
             | m_n(u,v)- G(u,H_k(v))|^2
             \PROB_{X_{n+1}}(du) \PROB_{(X_{n},...,X_{n-k+1})}(dv)
             \nonumber \\
             &&
             \leq
             (\log n)^3  \cdot
n^{- \frac{2 \cdot \min\{p_G,p_H\}}{2 \cdot \min\{p_G,p_H\}+(d+1)}}.
           \end{eqnarray}
           Let
           \[
m(x_{k+1},x_k,\dots,x_1)=\EXP\{Y_{k+1}|X_{k+1}=x_{k+1}, \dots,X_{1}=x_{1}\}
\]
be the regression function to $((X_{k+1},\dots,X_1),Y_{k+1})$.
By the assumptions on $(X_t,Y_t)$ we have
\[
m(x_{k+1},x_k,\dots,x_1)=G(x_{k+1},H_k(x_k,\dots,x_1))
\]
and
        \[
        m(x_{n+1},x_n,\dots,x_{n-(k-1)})=\EXP\{Y_{n+1}|X_{n+1}=x_{n+1}, \dots,X_{n-(k-1)}
        =x_{n-(k-1)}\},
\]
from which we can conclude by a standard decomposition of the
$L_2$ risk in nonparametric regression (cf., e.g., Section 1.1 in Gy\"orfi et
al. (2002))
           \begin{eqnarray*}
             &&
               \EXP \left\{
\left|
Y_{n+1}-m_n(X_{n+1},X_n, \dots, X_{n-(k-1)})
\right|^2
\right\}
\\
             &&
             =  \EXP \Bigg\{
\bigg|
(Y_{n+1}-m(X_{n+1},X_n, \dots, X_{n-(k-1)})
\\
&&
\hspace*{2cm}
+(m(X_{n+1},X_n, \dots, X_{n-(k-1)})
-m_n(X_{n+1},X_n, \dots, X_{n-(k-1)}))
\bigg|^2
\Bigg\}
\\
&&
=
               \EXP \left\{
\left|
Y_{n+1}-m(X_{n+1},X_n, \dots, X_{n-(k-1)}
\right|^2
\right\}
\\
&&
\hspace*{1cm}
+
\EXP \left\{ \left|
m(X_{n+1},X_n, \dots, X_{n-(k-1)})
-m_n(X_{n+1},X_n, \dots, X_{n-(k-1)})
\right|^2
\right\}
\\
&&
=
\min_{g: (\Rd)^{k+1} \rightarrow \R}
\EXP \left\{
\left|
Y_{n+1}-g(X_{n+1},X_n, \dots, X_{n-(k-1)})
\right|^2
\right\}
\\
&&
\hspace*{1cm}
+
             \EXP \int
             | m_n(u,v)- G(u,H_k(v))|^2
             \PROB_{X_{n+1}}(du) \PROB_{(X_{n},...,X_{n-k+1})}(dv).
             \end{eqnarray*}

                   {\it In the second step of the proof} we show
           \begin{eqnarray}
             \label{pth1eq2}
             &&
             \EXP \Bigg\{
             \int
             | m_n(u,v)- G(u,H_k(v))|^2
             \PROB_{X_{n+1}}(du) \PROB_{(X_{n},...,X_{n-k+1})}(dv)
             \nonumber \\
             &&
             \hspace*{2cm}
             -
             \frac{6 \cdot k+6}{n-k} \sum_{i=k+1}^n
              |m_n(X_i,X_{i-1},\dots,X_{i-k})-m(X_i,X_{i-1},\dots,X_{i-k})
             |^2
             \Bigg\}
             \nonumber \\
             &&
             \leq
c_{24} \cdot             (\log n)^6 \cdot
             n^{-\frac{ 2 \cdot \min\{p_G,p_H\}}{\min\{p_G,p_H\}+d+1}}
             \end{eqnarray}
           Set
           \begin{eqnarray*}
             &&
             T_n=
             \int
             | m_n(u,v)- G(u,H_k(v))|^2
             \PROB_{X_{n+1}}(du) \PROB_{(X_{n},...,X_{n-k+1})}(dv)
             \\
             &&
             \hspace*{0.5cm}
             -
             \frac{6 \cdot k+6}{n-k} \sum_{i=k+1}^n
              |m_n(X_i,X_{i-1},\dots,X_{i-k})-m(X_i,X_{i-1},\dots,X_{i-k})
             |^2.
             \end{eqnarray*}
           Then
           \begin{eqnarray*}
             &&
             T_n \leq
             \int
             | m_n(u,v)- G(u,H_k(v))|^2
             \PROB_{X_{n+1}}(du) \PROB_{(X_{n},...,X_{n-k+1})}(dv)
             \\
             &&
             \hspace*{0.1cm}
             -
             \frac{6 \cdot k + 6}{n-k} \sum_{i= k+1+l \cdot (k+1), \atop
               l \in \N_0, k+1+l \cdot (k+1) \leq n}
              |m_n(X_i,X_{i-1},\dots,X_{i-k})-m(X_i,X_{i-1},\dots,X_{i-k})
             |^2
             \\
             &&
             =
             \int
             | m_n(u,v)- G(u,H_k(v))|^2
             \PROB_{X_{n+1}}(du) \PROB_{(X_{n},...,X_{n-k+1})}(dv)
             \\
             &&
             \hspace*{0.5cm}
             -
             \frac{n_k \cdot (6 \cdot k + 6)}{3 \cdot (n-k)} \cdot \frac{3}{n_k}\sum_{i= k+1+l \cdot (k+1), \atop
               l \in \N_0, k+1+l \cdot (k+1) \leq n}
             \bigg|m_n(X_i,X_{i-1},\dots,X_{i-k})
             \\
             &&
             \hspace*{8cm}
             -m(X_i,X_{i-1},\dots,X_{i-k})
             \bigg|^2
           \end{eqnarray*}
           where
           \[
           n_k=| \{ l \in \N_0 \, : \,
           k+1+l \cdot (k+1) \leq n \} |
           \geq
           \left\lfloor \frac{n}{k+1} \right\rfloor
           \geq \frac{n}{2k+2}
           \]
           is the number of terms in the sum on the right-hand side
           above and consequently 
           \begin{eqnarray*}
             &&
             T_n
             \leq
             \int
             | m_n(u,v)- G(u,H_k(v))|^2
             \PROB_{X_{n+1}}(du) \PROB_{(X_{n},...,X_{n-k+1})}(dv)
             \\
             &&
             \hspace*{1cm}
             -
             \frac{3}{n_k}\sum_{i= k+1+l \cdot (k+1), \atop
               l \in \N_0, k+1+l \cdot (k+1) \leq n}
              |m_n(X_i,X_{i-1},\dots,X_{i-k})-m(X_i,X_{i-1},\dots,X_{i-k})
             |^2.
             \end{eqnarray*}
           Let
$\delta_n \geq c_{25}/n$.
           Then
           \[
\EXP\{ T_n\} \leq \delta_n + \int_{\delta_n}^{4 \beta_n^2} \PROB\{ T_n > t\} \, dt.
           \]
           We will apply Lemma \ref{le2} in order to bound
           $\PROB\{ T_n > t\}$ for $t>\delta_n$. Here we will
           replace $n$ by $n_k$ and $\F_n$ by
           $\F(k,K_{n,1},K_{n,2},L_{n,1},L_{n,2})$.
By Lemma \ref{le6}
we know for $u > c_{25} \cdot \delta_n/\beta_n^2$
\begin{eqnarray*}
  &&
  \log \Nu_2 \left( u , \left\{
(T_{\beta_n} f-m)^2 \, : \, f \in \F_n,
\frac{1}{n} \sum_{i=1}^n
|T_{\beta_n} f(x_i)-m(x_i)|^2 \leq \frac{\delta}{\beta_n^2}
\right\} , x_1^n \right)
\\
&&
\leq
  \log \Nu_2 \left( u , \left\{
(T_{\beta_n} f-m)^2 \, : \, f \in \F_n
\right\} , x_1^n \right)
\\
&&
\leq
  \log \Nu_2 \left( \frac{u}{4 \beta_n} , \left\{
T_{\beta_n} f \, : \, f \in \F_n
\right\} , x_1^n \right)
\\
&&
\leq
c_{26} \cdot k \cdot (\max\{L_{n,1}, L_{n,2}\})^2
\cdot (\max\{K_{n,1}, K_{n,2}\})^2 \cdot (\log n)^2.
    \\
    &&
    \leq
    c_{27} \cdot n^{\frac{(d+1)}{2 \cdot \min\{p_G,p_H\}+d+1}} \cdot (\log n)^2.
  \end{eqnarray*}
Consequently (\ref{le2eq1}) is satisfied for
\[
\delta_n=c_{28} \cdot \frac{n^{\frac{d+1}{2 \cdot \min\{p_G,p_H\}+d+1}} \cdot (\log n)^6}{n}
=
c_{28} \cdot (\log n)^6  \cdot n^{-\frac{2 \cdot \min\{p_G,p_H\}}{2 \cdot \min\{p_G,p_H\}+d+1}}.
\]
Application of Lemma \ref{le2} yields
\[
\EXP\{ T_n\} \leq \delta_n + c_{29} \cdot \frac{\beta_n^2}{n} \cdot
\exp\left( - \frac{n \cdot \delta_n}{c_{14} \cdot \beta_n^2}\right),
\]
which implies (\ref{pth1eq2}).

{\it In the third step of the proof} we show
\begin{eqnarray}
  && \label{pth1eq3}
  \EXP \left\{
             \frac{1}{n-k} \sum_{i=k+1}^n
              |m_n(X_i,X_{i-1},\dots,X_{i-k})-m(X_i,X_{i-1},\dots,X_{i-k})
             |^2
             \right\}
             \nonumber \\
             &&
             \leq
             c_{30} \cdot             (\log n)^6 \cdot
             n^{-\frac{ 2 \cdot \min\{p_G,p_H\}}{\min\{p_G,p_H\}+d+1}}.
  \end{eqnarray}
To do this, we set
\[
\bar{T}_n=
\frac{1}{n-k} \sum_{i=k+1}^n
              |m_n(X_i,X_{i-1},\dots,X_{i-k})-m(X_i,X_{i-1},\dots,X_{i-k})
             |^2
\]
and define $\delta_n$ as in the second step of the proof
(for $c_{28}$ sufficiently large).
Then
\[
\EXP\{ \bar{T}_n \}
\leq \delta_n + \int_{\delta_n}^{4 \beta_n^2} \PROB\{ \bar{T}_n > t\} \, dt.
\]
To bound $ \PROB\{ \bar{T}_n > t\}$ for $t \geq \delta_n$,
we apply Lemma \ref{le1} conditioned on $X_1, \dots, X_n$
and with sample size $n-k$ instead of $n$
and with
           $\F(k,K_{n,1},K_{n,2},L_{n,1},L_{n,2})$
instead of $\F_n$. As in the proof of
the second step we see that (\ref{le1eq1}) holds for $\delta \geq
\delta_n/12$. Furthermore, we get by application of
Lemma \ref{le3} b) and
Lemma \ref{le5}
\begin{eqnarray*}
  &&
  \min_{f \in \F(k,K_{n,1},K_{n,2},L_{n,1},L_{n,2})}
\frac{1}{n-k} \sum_{i=k+1}^n
              |f(X_i,X_{i-1},\dots,X_{i-k})-m(X_i,X_{i-1},\dots,X_{i-k})
             |^2
             \\
             &&
             \leq
             \left(
             \min_{f \in \F(k,K_{n,1},K_{n,2},L_{n,1},L_{n,2})} \|f-m\|_{\infty, [0,1]^d}
             \right)^2
             \\
             &&
             \leq
             c_{31} \cdot \max\{L_{n,1}^{-\frac{2 p_H}{d+1}},
             L_{n,2}^{-\frac{2 p_G}{d+1}} \}
             \leq
             c_{28}/2 \cdot             (\log n)^6 \cdot
             n^{-\frac{ 2 \cdot \min\{p_G,p_H\}}{\min\{p_G,p_H\}+d+1}}=\delta_n/2.
\end{eqnarray*}
Consequently, we get by Lemma \ref{le1}
\begin{eqnarray*}
  \EXP\{ \bar{T}_n \}
&\leq& \delta_n + \int_{\delta_n}^{4 \beta_n^2} \PROB\{ \bar{T}_n > t\} \, dt.
  \\
  &\leq& \delta_n +4 \beta_n^2 \cdot
  \PROB\{ \bar{T}_n > \delta_n/2 + \delta_n/2\}
  \\
  &\leq&
  \delta_n +4 \beta_n^2 \cdot c_{32} \cdot \exp( - c_{33} \cdot (n-k) \cdot
  \frac{\delta_n}{2}) + 4 \beta_n^2 \cdot \frac{c_{34}}{n},
  \end{eqnarray*}
which implies (\ref{pth1eq3}).

{\it In the fourth step of the proof} we conclude the proof of
Theorem \ref{th1} by showing (\ref{pth1eq1}). Define $T_n$ and
$\bar{T}_n$ as in the second and in the third step of
the proof, resp. Then (\ref{pth1eq2}) and (\ref{pth1eq3})
imply
\begin{eqnarray*}
             &&
             \EXP \int
             | m_n(u,v)- G(u,H_k(v))|^2
             \PROB_{X_{n+1}}(du) \PROB_{(X_{n},...,X_{n-k+1})}(dv)
             \\
             &&
             \leq \EXP\{ T_{n,1} \}
             +
             (6 \cdot k + 6) \cdot
             \EXP\{ T_{n,2} \}
             \leq
             c_{35} \cdot             (\log n)^6 \cdot
             n^{-\frac{ 2 \cdot \min\{p_G,p_H\}}{\min\{p_G,p_H\}+d+1}}.
  \end{eqnarray*}
\hfill $\Box$

\newpage
\begin{appendix}
  \section{Proof of Lemma \ref{le1}}
 By the Markov inequality and assumption
 (\ref{se52eq1}) we have
 \begin{eqnarray*}
   &&
   \PROB \left\{
\exists i \in \{1, \dots, n\}: |Y_i| > \beta_n
\right\}
\\
&&
\leq
   \PROB \left\{
\exists i \in \{1, \dots, n\}: |W_i| > \beta_n/2
\right\}
\\
&&
\leq
n \cdot \max_{i=1, \dots, n}
   \PROB \left\{
\exp( |W_i|^2/K^2) > \exp((\beta_n/2)^2/K^2)
\right\}
\\
&&
\leq
\frac{n}{\exp( (\beta_n)^2 /(4 K^2))}
\cdot
\left(
1 + \frac{\sigma_0^2}{K^2}
\right).
   \end{eqnarray*}
 This together with Lemma 1 in Kohler and Krzy\.zak (2020)
 implies
 \begin{eqnarray*}
   &&
  \PROB \Bigg\{
  \frac{1}{n} \sum_{i=1}^n | m_n(x_i)-m(x_i)|^2
>
3 \cdot \left(
\delta_n
+
\min_{f \in \F_n}
  \frac{1}{n} \sum_{i=1}^n | f(x_i)-m(x_i)|^2
\right)
\Bigg\}
\\
&&
\leq
  \PROB \Bigg\{
  \frac{1}{n} \sum_{i=1}^n | m_n(x_i)-m(x_i)|^2
>
3 \cdot \left(
\delta_n
+
\min_{f \in \F_n}
  \frac{1}{n} \sum_{i=1}^n | f(x_i)-m(x_i)|^2
  \right),
  \\
  &&
  \hspace*{8cm}
  \max_{i=1, \dots, n}|Y_i| \leq \beta_n
\Bigg\}
+
\frac{c_{36}}{n}
  \\
  &&
  \leq
  \PROB \Bigg\{
  \frac{1}{n} \sum_{i=1}^n (m_n(x_i)-m_n^*(x_i)) \cdot W_i
  \geq
  \frac{1}{24} \cdot
  \frac{1}{n} \sum_{i=1}^n | m_n(x_i)-m_n^*(x_i)|^2 + \frac{\delta_n}{2}
\Bigg\}
+
\frac{c_{36}}{n},
   \end{eqnarray*}
 where we have set
 \[
 m_n^* = \arg \min_{f \in \F_n}
  \frac{1}{n} \sum_{i=1}^n | f(x_i)-m(x_i)|^2.
 \]
 So it remains to show
 \begin{eqnarray*}
   &&
  \PROB \Bigg\{
  \frac{24}{n} \sum_{i=1}^n (m_n(x_i)-m_n^*(x_i)) \cdot W_i
  \geq
  \frac{1}{n} \sum_{i=1}^n | m_n(x_i)-m_n^*(x_i)|^2 + 12 \cdot \delta_n
  \Bigg\}
  \\
  &&
\leq
c_{37} \cdot \exp \left(
- \frac{n \cdot \min \{ \delta_n, \sigma_0^2 \} }{c_{37}}
\right),
 \end{eqnarray*}
 which we do next.

 For $f:\Rd \rightarrow \R$ set
\[
\|f\|_n^2
=
\frac{1}{n} \sum_{i=1}^n |f(X_i)|^2.
\]
 We have
\begin{eqnarray*}
&&
\PROB \left\{
\|m_n - m_n^* \|_n^2 + 12 \cdot \delta_n
\leq
\frac{24}{n}
\sum_{i=1}^n ( m_n(x_i)-m_n^* (x_i)) \cdot W_i
\right\}
\leq P_1 + P_2
\end{eqnarray*}
where
\[
P_1 =
\PROB \left\{
\frac{1}{n} \sum_{i=1}^n W_i^2 > 2 \sigma_0^2
\right\}
\]
and
\begin{eqnarray*}
&&
P_2 =
\PROB \left\{
\frac{1}{n} \sum_{i=1}^n W_i^2 \leq 2 \sigma_0^2,
\|m_n - m_n^* \|_n^2 + 12 \cdot \delta_n
\leq
\frac{24}{n}
\sum_{i=1}^n ( m_n(x_i)-m_n^* (x_i)) \cdot W_i
\right\}.
\end{eqnarray*}
 By the Markov inequality and assumption
 (\ref{se52eq1}) we have
\begin{eqnarray*}
P_1
&=& \PROB\left\{
\sum_{i=1}^n W_i^2 / K^2 > 2 n \sigma_0^2 /K^2
\right\}
\\
&\leq&
\PROB\left\{
\exp \left( \sum_{i=1}^n W_i^2 / K^2 \right)
>
\exp \left( 2 n \sigma_0^2 /K^2 \right)
\right\}
\\
&\leq&
\exp \left(
- 2 n \sigma_0^2 /K^2
\right)
\cdot
\EXP
\left\{
\exp (
\sum_{i=1}^n W_i^2 / K^2
)
\right\}
\\
&\leq&
\exp \left(
- 2 n \sigma_0^2 /K^2
\right)
\cdot
\left( 1 + \sigma_0^2/K^2 \right)^n
\\
&\leq&
\exp \left(
- 2 n \sigma_0^2 /K^2
\right)
\cdot
\exp \left(
n \cdot \sigma_0^2 /K^2
\right)
=
\exp \left(
-  n \sigma_0^2 /K^2
\right)
.
\end{eqnarray*}

To bound $P_2$, we observe first that
$1/n \sum_{i=1}^n W_i^2 \leq 2 \sigma_0^2$
together with the Cauchy-Schwarz inequality implies
\begin{eqnarray*}
\frac{24}{n}
\sum_{i=1}^n ( m_n(x_i)-m_n^* (x_i)) \cdot W_i
& \leq &
24
\cdot
\sqrt{
\frac{1}{n}
\sum_{i=1}^n ( m_n(x_i)-m_n^* (x_i))^2
}
\cdot \sqrt{2 \sigma_0^2}
\end{eqnarray*}
hence inside of $P_2$ we have
\[
\frac{1}{n}
\sum_{i=1}^n (  m_n(x_i)-m_n^* (x_i))^2
\leq
1152 \sigma_0^2.
\]
Set
\[
S= \min \{ s \in \mathbb N_0 \, : \, 4 \cdot 2^s \delta_n > 1152 \sigma_0^2 \}.
\]
Application of the peeling device (cf. Section 5.3 in van de Geer (2000))
yields
\begin{eqnarray*}
P_2
&&
=
\sum_{s=1}^S
\PROB \Bigg\{\frac{1}{n} \sum_{i=1}^n W_i^2 \leq 2 \sigma_0^2,
12 \cdot 2^{s-1} \delta_n \cdot I_{\{s \neq 1\}}
\leq
\|m_n - m_n^* \|_n^2  <12 \cdot 2^s \delta_n,
\\
&&
\hspace*{2cm}
\|m_n - m_n^* \|_n^2 + 12 \delta_n
\leq
\frac{24}{n}
\sum_{i=1}^n ( m_n(x_i)-m_n^* (x_i)) \cdot W_i
\Bigg\}
\\
&&
\leq
\sum_{s=1}^S
\PROB \Bigg\{
\frac{1}{n} \sum_{i=1}^n W_i^2 \leq 2 \sigma_0^2,
\|m_n - m_n^* \|_n^2 <12 \cdot 2^s \delta_n,
\\
&&
\quad\quad\quad
\quad\quad\quad
\frac{1}{2} \cdot 2^{s} \delta_n
\leq
\frac{1}{n}
\sum_{i=1}^n ( m_n(x_i)-m_n^* (x_i)) \cdot W_i
\Bigg\}
\end{eqnarray*}
The probabilities in the above sum can be bounded by
Corollary 8.3 in van de Geer (2000) (use there
$R= \sqrt{12 \cdot 2^s \delta_n}$,
$\delta= \frac{1}{2} \cdot 2^s \delta_n$
and $\sigma= \sqrt{2} \sigma_0$). This yields
\begin{eqnarray*}
&&
P_2
\leq
\sum_{s=1}^\infty
c_{38} \cdot
\exp
\left(
-
\frac{
n \cdot
(
\frac{1}{2} \cdot 2^{s} \delta_n
)^2
}{
4 c_{60} \cdot 12 \cdot 2^s \delta_n
}
\right)
=
\sum_{s=1}^\infty
c_{38} \cdot
\exp
\left(
-
\frac{
n \cdot 2^s \cdot \delta_n
}{
c_{38}
}
\right)
\\
&&
\leq
\sum_{s=1}^\infty
c_{38} \cdot
\exp
\left(
-
\frac{
n \cdot (s+1) \cdot \delta_n
}{
c_{38}
}
\right)
\leq
c_{39} \cdot
\exp
\left(
-
\frac{
n \cdot \delta_n
}{
c_{39}
}
\right).
\end{eqnarray*}

  \quad \hfill $\Box$

  \section{Proof of Lemma \ref{le2}}
For $f:\Rd \rightarrow \R$ set
\[
\|f\|_n^2
=
\frac{1}{n} \sum_{i=1}^n |f(X_i)|^2.
\]
    We have
\begin{eqnarray*}
  &&
  \PROB \left\{
    \int |m_n(x)-m(x)|^2 \PROB_X(dx)
>
\delta_n
+
3
\frac{1}{n}
\sum_{i=1}^n
|m_n(X_i)-m(X_i)|^2
\right\}
\\
&&=
\PROB \Bigg\{
2 \int |m_n(x)-m(x)|^2 \PROB_X (dx) - 2 \| m_n-m\|_n^2
\\
&&
\hspace*{2cm}
>
\delta_n
+
\int |m_n(x)-m(x)|^2 \PROB_X (dx)
+
\| m_n-m\|_n^2
\Bigg\}
\\
&&\leq
\PROB
\Bigg\{
\exists f \in \F_n:
\frac{
\left|
\int |T_{\beta_n} f (x)-m(x)|^2 \PROB_X (dx) -  \| T_{\beta_n} f-m\|_n^2
\right|
}{
\delta_n
+
\int |T_{\beta_n} f (x)-m(x)|^2 \PROB_X (dx)
+  \| T_{\beta_n} f-m\|_n^2
}
> \frac{1}{2}
\Bigg\}
.
\end{eqnarray*}
The probability above can be bounded by Theorem 19.2
in Gy\"orfi et al. (2002) (which we apply with
\[
\F = \left\{
(T_{\beta_n} f - m)^2 \, : \,
f \in \F_n
\right\},
\]
$K=4 \beta_n^2$, $\epsilon=1/2$, and $\alpha=\delta_n$.)
This yields
\[
P_{1,n}
\leq
15 \cdot
\exp \left(
- \frac{n \cdot \delta_n }{c_{40} \cdot \beta_n^2}
\right)
.
\]
\quad \hfill $\Box$

  \end{appendix}


\begin{thebibliography}{*}


\bibitem{BHLM19}
  Bartlett, P. L., Harvey, N., Liaw, C., and Mehrabian, A. (2019).
  Nearly-tight VC-dimension bounds for piecewise linear neural networks.
  {\it Journal of Machine Learning Research}, {\bf 20}, pp. 1--17.

\bibitem{BK17}
Bauer, B., and Kohler, M. (2019).
On deep learning as a remedy for the curse of dimensionality in nonparametric regression. {\it Annals of Statistics}, {\bf 47}, pp. 2261-2285.

\bibitem{Bru90}
Bruck, J. (1990). On the convergence properties of the Hopfield model. {\it Proceedings of the IEEE}, {\bf 78}, pp.1579-1585.

\bibitem{ChMeGuBaBoScBe14}
Cho, K., Van Merri\"enboer, B., Gulcehre, C., Bahdanau, D., Bougares, F., Schwenk, H., and Bengio, Y. (2014).
Learning phrase representations using RNN encoder-decoder for statistical machine translation.
arXiv:1406.1078.


\bibitem{ElSh16}
Eldan, R., and Shamir, O. (2016).
The Power of Depth for Feedforward Neural Networks.
{\it Proc. Mach Learn. Res. (PMLR)}, \textbf{49},
pp. 907--940.

\bibitem{vdG00}
  van de Geer, S. (2000).
  {\it Empirical Processes in M-Estimation}.
  Cambridge Series in Statistical and Probabilistic Mathematics,
  Cambridge University Press.



\bibitem{GrLiFeBeBuSc08}
Graves, A., Liwicki, M., Fernandez, S., Bertolami, R., Bunke, H. and Schmidhuber, J. (2008).
A novel connectionist system for unconstrained handwriting recognition.
{\it IEEE Transactions on Pattern Analysis and Machine Intelligence}, {\bf 31}, pp. 855-868.


\bibitem{GrMoHi13}
Graves, A., Mohamed, A.R., and Hinton, G. (2013).
Speech recognition with deep recurrent neural networks.
\textit{2013 IEEE international Conference on Acoustics, Speech and Signal Processing (ICASSP'2013)}, pp. 6645-6649.

  
\bibitem{GrSc05}
Graves, A. and Schmidhuber, J. (2005). Framewise phoneme classification with bidirectional LSTM and other neural network architectures.
{\it Neural Networks}, {\bf 18}, pp. 602-610.

\bibitem{GrSc09}
Graves, A. and Schmidhuber, J. (2009).
Offline handwriting recognition with multidimensional recurrent neural networks.
\textit{Advances in Neural Information Processing Systems (NIPS'2009)}, pp.  545-552.


 \bibitem{GKKW02}
 Gy\"orfi, L., Kohler, M., Krzy\.zak, A., and Walk, H. (2002).
 {\it A Distribution--Free Theory of Nonparametric Regression}.
 Springer.

\bibitem{HBB20}
  Hewamalage, H., Bergmeir, C., and Bandara, K. (2020).
  Recurrent neural networks for time series forecasting:
  current status and future directions.
To appear in  {\it International Journal of Forecasting}.

\bibitem{HoSc97}
Hochreiter, S. and Schmidhuber, J. (1997).
Long short-term memory.
{\it Neural Computation}, {\bf 9}, pp.1735-1780.


\bibitem{KoLa18}
Kohler, M., and Langer, S. (2020).
On the rate of convergence of fully connected deep neural network regression estimates. To appear in {\it Annals of Statistics} 2021,
ArXiv:1908.11133.

\bibitem{KoKr15}
Kohler, M., and Krzy\.zak, A. (2017).
Nonparametric regression based on hierarchical interaction models.
\textit{IEEE Transaction on Information Theory}, \textbf{63}, pp. 1620-1630.

\bibitem{KoKr20}
Kohler, M. and Krzy\.zak, A. (2020).
Estimation of a density using an improved surrogate model.
Submitted for publication.

\bibitem{Kos88}
Kosko, B. (1988).
Bidirectional associative memories.
{\it IEEE Transactions on Systems, Man and Cybernetics}, {\bf 18}, pp. 49-60.

\bibitem{KrMaKrPeBaKeGoBeCoPa16}
Krueger, D., Maharaj, T., Kramár, J., Pezeshki, M., Ballas, N., Ke, N.R., Goyal, A., Bengio, Y., Courville, A. and Pal, C. (2016). Zoneout: Regularizing rnns by randomly preserving hidden activations. ArXiv:1606.01305.

\bibitem{LuShYaZh20}
Lu, J., Shen, Z., Yang, H., and Zhang, S. (2020).
Deep network approximation for smooth functions.
ArXiv:2001.03040

\bibitem{MaCa20}
Mas, A. and Carre, C. (2020).
Prediction of Hilbertian autoregressive processes : a Recurrent Neural
Network approach.
ArXiv:2008.11155.

\bibitem{MaSpAs18}
Makridakis, S, Spiliotis, E., and Assimakopolulos, V. (2018)
Statistical and machine learning forecasting methods: Concerns and ways forward.
\textit{PLOS One}, \textbf{13}, pp. 1-26.

\bibitem{PeSoMa14}
Pennington, J., Socher, R., and Manning, C. D. (2014).
Glove: Global vectors for word representation.
\textit{Proceedings of the 2014 conference on empirical methods in natural language processing (EMNLP)}, pp. 1532-1543.

\bibitem{RaWa17}
Rawat, W., and Wang, Z. (2017).
Deep convolutional neural networks for image classification:
a comprehensive review.
{\it Neural Computation}, {\bf 29}, pp. 2352-2449.


\bibitem{SaSaBaCoVa18}
Salehinejad, H., Sankar, S., Barfett, J., Colak, E., and Valaee, S. (2018). Recent advances in recurrent neural networks.
arXiv preprint arXiv:1801.01078.

\bibitem{Sch15}
Schmidhuber, J. (2015).
Deep learning in neural networks: an overview.
\textit{Neural Networks}, \textbf{61}, pp. 85-117.



\bibitem{Sch17}
Schmidt-Hieber, J. (2020).
Nonparametric regression using deep neural networks with ReLU activation function (with discussion). {\it Annals of Statistics}, \textbf{48}, pp. 1875-1897 .


\bibitem{Smy20}
Smyl, S. (2020). A hybrid method of exponential smoothing and
recurrent neural networks for time series forecasting.
{\it International Journal of Forecasting},
\textbf{36}, pp. 75-85.

\bibitem{SoLINgMa11}
Socher, R., Lin, C.C.Y., Ng, A.Y., and Manning, C.D. (2011).
 Parsing natural scenes and natural language with recursive neural networks.  {\it Proceedings of the International Conference on Machine Learning (ICML'2011)}.

\bibitem{SrHiKrSuSa14}
Srivastava, N., Hinton, G., Krizhevsky, A., Sutskever, I., and Salakhutdinov, R. (2014). Dropout: a simple way to prevent neural networks from overfitting.  {\it Journal of Machine Learning Research}, {\bf 15}, pp. 1929-1958.

\bibitem{Sto82}
Stone, C. J. (1982).
Optimal global rates of convergence for nonparametric regression.
\textit{Annals of Statistics}, \textbf{10}, pp. 1040-1053.

\bibitem{Ya18}
  Yarotsky, D. (2018).
  Optimal approximation of continuous functions by very deep ReLU networks.
  {\it COLT}, \textbf{75}, pp. 639-649.

\bibitem{YaZh19}
Yarotsky, D., and Zhevnerchuk, A. (2019).
The phase diagram of approximation rates for deep
neural networks.
arXiv:1906.09477.

\end{thebibliography}
\end{document}